\documentclass[10pt,twocolumn,letterpaper]{article}

\usepackage{cvpr}
\usepackage{times}
\usepackage{epsfig}
\usepackage{graphicx}
\usepackage{amsmath}
\usepackage{amssymb}

\usepackage{booktabs}   
\usepackage{lipsum}
\usepackage{booktabs}
\usepackage{multirow}
\usepackage{graphicx} % for '\resizebox` macro
\usepackage{tabularx} % for 'tabularx' environment

\usepackage[breaklinks=true,bookmarks=false]{hyperref}

\cvprfinalcopy % *** Uncomment this line for the final submission

\def\cvprPaperID{****} % *** Enter the CVPR Paper ID here

\begin{document}

%%%%%%%%% TITLE
\title{Group Ensemble: Learning an Ensemble of ConvNets in a single ConvNet}

\author{Hao Chen\\
University of Maryland, College Park\\
{\tt\small chenh@cs.umd.edu}
\and
Abhinav Shrivastava \\
University of Maryland, College Park\\
{\tt\small abhinav@cs.umd.edu }
}

\maketitle
%\thispagestyle{empty}

%%%%%%%%% ABSTRACT
\begin{abstract} \label{sec:intro}
Ensemble learning is a general technique to improve accuracy in machine learning.
However, the heavy computation of a ConvNets ensemble limits its usage in deep learning.
In this paper, we present Group Ensemble Network (GENet), an architecture incorporating an ensemble of ConvNets in a single ConvNet.
Through a shared-base and multi-head structure, GENet is divided into several groups to make explicit ensemble learning possible in a single ConvNet.
Owing to group convolution and the shared-base, \emph{GENet can fully leverage the advantage of explicit ensemble learning while retaining the same computation as a single ConvNet}.
Additionally, we present Group Averaging, Group Wagging and Group Boosting as three different strategies to aggregate these ensemble members.
Finally, GENet outperforms larger single networks, standard ensembles of smaller networks, and other recent state-of-the-art methods on CIFAR and ImageNet.
Specifically, group ensemble reduces the top-1 error by 
1.83\% for ResNeXt-50 on ImageNet.
We also demonstrate its effectiveness on action recognition and object detection tasks.
\end{abstract}

%%%%%%%%% BODY TEXT

\section{Introduction}
Ensemble learning, a general technique of combining several models to create a more accurate one, has a vast and varied history in machine learning. The goal of different ensemble methods is to either lower the variance or bias of the final model, or its error rates. These methods range from the simple committee of experts and ad-hoc averaging to techniques such as bagging, boosting, stacking, etc. In the case of deep neural networks, the models of choice for recognition, ensembling strategies can be grouped into explicit and implicit methods.

\begin{figure}[t!]
    \centering
    \includegraphics[width=\linewidth]{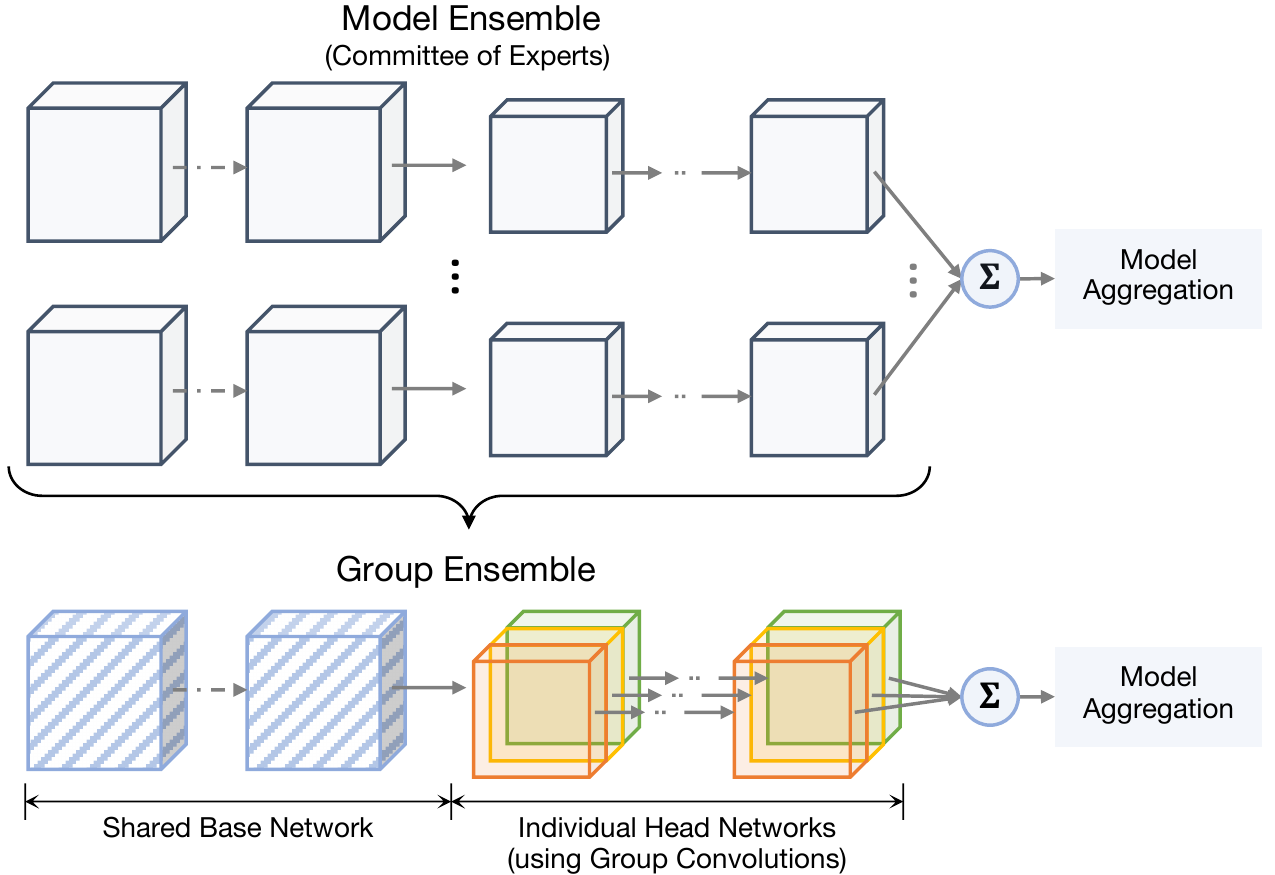}
    \vspace{-0.3em}
    \caption{\textbf{Framework of Model Ensemble \vs Group Ensemble}.
\textbf{Top}: Model Ensemble: ensemble members are trained separately in multiple ConvNets.
\textbf{Bottom}: Group Ensemble: ensemble members are incorporated in a single ConvNet.}
    \label{fig:group-ens-idea}
\vspace{-.5em}
\end{figure}

Explicit ensembling strategies are direct extensions of popular ensemble methods; for example, a committee of experts~\cite{zhou2012ensemble} -- a mainstay in recognition competitions, boosting in neural networks~\cite{drucker1993improving,schwenk2000boosting,schwenk1997adaboosting,moghimi2016boosted,kuznetsov2014multi,cortes2014deep}. However, since deep neural networks are already computationally expensive, explicit ensembling additionally requires substantial computational resources compared to individual networks. 
On the contrary, implicit strategies focus on learning a single model with ensemble-like properties. 
These implicit strategies follow two complementary directions: imitating ensemble in a single model using stochastic operations (\eg.,
DropOut~\cite{Srivastava2014DropoutAS},
DropConnect~\cite{wan2013regularization}, DropBlock~\cite{Ghiasi2018DropBlockAR}, StochDepth~\cite{huang2016deep},  Shake-Shake~\cite{Gastaldi17ShakeShake}), and encoding ensemble architecturally using `multiple paths' (\eg., ResNet~\cite{he2016deep}, ResNeXt~\cite{xie2017aggregated}, DenseNet~\cite{Huang2016DenselyCC}, Inception series~\cite{Szegedy_2015,Szegedy_2016,Szegedy2016Inceptionv4IA}).
% and their re-interpretations~\cite{lv2019deep}). 
However, generally, these implicit approaches cannot take advantage of different model aggregation strategies (\eg., bagging, boosting, stacking), which are key in ensemble learning. 
More importantly,  these implicit ensemble approaches can be enhanced further by explicit ensemble learning.
In this paper, we propose a simple, efficient, and effective approach at the intersection of these two strategies. Our approach \emph{explicitly} learns an ensemble, in a \emph{single model} itself, formulated using widely used \emph{architectural recipes}. 

ConvNets are a special class of models, which learn a bottom-up, feedforward hierarchy of feature representations. Several recent studies~\cite{Morcos2018InsightsOR,Kornblith2019SimilarityON,Raghu2017SVCCASV} presented empirical insights that different instantiations of the same ConvNet architecture (either with different random initialization or trained on different splits of data) converge to similar solutions in lower layers (\eg., representing low-level filters). Moreover, Raghu et al.~\cite{Raghu2017SVCCASV}, compared a VGG-style ConvNet with a ResNet and demonstrated that even different ConvNet architectures learn similar lower layers. This insight leads to the key design principle of our method -- lower layers of different members of a ConvNet ensemble can be shared, thus reducing the computation cost. Next, we note that to utilize different model aggregation techniques from ensemble learning, higher layers and the prediction layer of individual members of a CovnNet ensemble should be independent. 

Based on these insights, we propose a shared base, multi-head architecture for learning an ensemble of ConvNets (illustrated in Figure~\ref{fig:group-ens-idea}). The multi-head architecture can be implemented efficiently using grouped convolutions~\cite{krizhevsky2012imagenet,xie2017aggregated}, where each group has a different classifier head. 
Finally, the outputs of the individual classifiers can be combined using any model aggregation technique. 
This simple formulation of a ConvNets ensemble in a single ConvNet allows for independent (\eg.,  wagging), sequential (\eg., boosting), and posthoc (\eg., averaging, stacking) ensembling in the same framework of end-to-end training of a ConvNet with minibatch Stochastic Gradient Descent.

We refer to our proposed network as \textbf{Group Ensemble Network}  (GENet), since a group is the constituent member of our ensemble. 
Owing to group convolution and parameter sharing,
\emph{GENet fully leverages the advantages of explicit ensemble learning, while retaining the same computation as a single ConvNet.}
Additionally, we present Group  Averaging,  Group  Wagging  and Group Boosting as three different strategies to aggregate these ensemble members.
Finally, compared with standard explicit ensemble learning, our GENEt achieves comparable performance while using much fewer resources (\eg, number of parameters, FLOPs, and memory consumption).
And with roughly similar parameters, GENet outperforms larger single networks, the standard ensemble of smaller networks, and other recent state-of-the-art methods on CIFAR and ImageNet.
Specifically, GENet reduces the top-1 error by 1.73\%/1.83\% for ResNet-50/ResNeXt-50 respectively on ImageNet.
We also demonstrate its effectiveness on action recognition and object detection tasks.

\section{Related Work}
\noindent{\textbf{Model Ensemble.}}
Ensemble learning has a rich and diverse history in machine learning~\cite{zhou2012ensemble,friedman2001elements}. By combining several well-trained models, ensemble learning~\cite{Rokach2009EnsemblebasedC,Caruana2004EnsembleSF} is an effective way to boost performance and is widely adopted in recognition competitions such as \cite{lin2014microsoft,Russakovsky_2015}.
Methods to build ensembles can be divided into two types: independent frameworks and dependent frameworks. For independent frameworks, each model is built independently, and their outputs are combined as the final output. Examples include: bagging (bootstrap aggregating) \cite{Breiman1996BaggingP}, wagging \cite{Bauer1997AnEC}, random forest ensemble~\cite{Breiman2001RandomF}, or simply a committee of experts~\cite{kittler1998combining,alpaydin1998techniques,perrone1994general}. Wagging (weight aggregation)~\cite{bauer1999empirical} is a variant of Bagging, which repeatedly perturbs the training set by adding Gaussian noise to sample weights. On the other hand, dependent frameworks use the output of a classifier in the construction of the next classifier; \eg., incremental batch learning, boosting \cite{Freund1996ExperimentsWA,Chen2019EmbeddingCD,cortes2014deep,kuznetsov2014multi,Peng_2016,Tolstikhin2017AdaGANBG,drucker1993improving}.
As opposed to these approaches, where ensemble members are trained separately in several networks or come from complex model families, members of our group ensemble are trained in a single ConvNet, which significantly reduce the model complexity and computation cost.

\medskip
\noindent{\textbf{Parameter Sharing.}}
Parameter sharing \cite{bromley1994signature,Goodfellow-et-al-2016,lee2015m,song2018collaborative,li2019ensemblenet} can be seen as one way to regularize parameters by requiring sets of parameters to be shared across several networks. It is a common practice to improve accuracy while maintaining the model size, which is widely used in neural architecture search (NAS) \cite{Pham2018EfficientNA} and multi-task learning (MTL) \cite{misra2016cross,ruder122019latent,meyerson2017beyond,long2015learning,rosenbaum2017routing}.
For MTL with convolution networks, several related tasks are optimized jointly~\cite{misra2016cross}, where some feature representations (early layers in deep ConvNet) are shared amongst these tasks; \eg action recognition and pose estimation~\cite{Gkioxari2014RCNNsFP}, object detection and instance segmentation~\cite{He_2017}, segmentation and surface normal estimation~\cite{misra2016cross}. 
Therefore, the network can be divided into two parts based on its parameters: shared network with generic parameters (generally bottom layers) and task-specific networks with separate parameters (generally top layers).
By leveraging training examples of several tasks, shared parameters are more regularized, which often leads to better generalization~\cite{caruana1997multitask}. 
Similar to multi-task learning, the parameters of the bottom layers are shared among multiple head classifiers in our group ensemble.
However, these classifiers focus on the same task, as opposed to MTL where they focus on different tasks.
Although trained on the same task, parameter sharing  still adds an extra regularization for shared-base parameters.

\medskip
\noindent{\textbf{Grouped Convolution.}}
AlexNet~\cite{krizhevsky2012imagenet} first introduced grouped convolution as an implementation convenience to distribute a ConvNet model over two GPUs. Then, ResNeXt~\cite{xie2017aggregated} further investigated the trade-off between groups and channel width for ConvNets, and concluded that adding more groups can improve model accuracy under given computation cost. Recently depth-wise convolution has been widely studied in many papers, like MobileNet~\cite{howard2017mobilenets}, Xception~\cite{chollet2017xception}, CSN~\cite{Tran2019VideoCW}. By introducing grouped convolution, all these methods are attempting to reduce the model size and computational cost, while maintaining or improving the model performance.
However, for the proposed group ensemble, it is leveraged to embed multiple ConvNets in one single network, by splitting the network into several branches using groups and treating them as individual members in an ensemble. Therefore, our approach benefits from explicit ensemble learning while keeping the computation complexity similar to a single ConvNet.

The work most related to ours are ~\cite{song2018collaborative,lee2015m,li2019ensemblenet, zhu2018knowledge}, which also utilizes the multi-head structure for ensemble learning. However,~\cite{song2018collaborative,lee2015m, zhu2018knowledge} do not consider computational efficiency and utilize heavy neural network architectures~(simply adding more child networks on the shared-base), while we reduce the head size to save computation resources.~\cite{song2018collaborative,li2019ensemblenet} focus only on the collaborative learning objective while we explore extensively for architecture design~(\eg, group numbers, split layers) and aggregation strategies.

\section{Group Ensemble}
To incorporate an ensemble of ConvNets in a single ConvNet, we propose Group Ensemble network~(GENet), where one network is divided into several groups~(as illustrated in Figure~\ref{fig:group-ens-idea}). 
The main idea behind this architecture is that bottom layers in a ConvNet are learning basic visual patterns (\eg, edges, colors, simple shapes), which can be shared among ensemble members. 
The multi-head structure can be efficiently implemented by group convolution where each group is one constituent member of our ensemble and has its own independent head classifier.
Besides, we present Group  Averaging,  Group  Wagging  and Group Boosting as three different strategies to aggregate these head classifiers.
Through group ensemble framework, GENet fully leverages the advantage of explicit ensemble learning while retaining the computation cost.

\subsection{Network Architecture} 
\label{sec:net-archi}
As shown in Figure~\ref{fig:group-ens-idea},
compared to model ensemble methods, GENet incorporates multiple members in a single ConvNet, thus benefiting from explicit ensemble learning while retaining the computation cost of a single ConvNet.
To construct multiple constituent members, we build a multi-head architecture on the shared-base, which can be efficiently implemented using grouped convolution.
Based on the shared feature representation, each head classifier generates its own high-level feature representations for the target task.
These independent heads introduce diversity among ensemble members and incorporate ensemble learning explicitly in a single ConvNet. 

Each head classifier is trained independently (the whole ensemble is still trained end-to-end simultaneously) with its own objective function,
\begin{equation}
\label{equ:ens-loss}
    Loss = \sum_{m=1}^{n} Loss_{m},
\end{equation}
where $Loss$ is the total objective function, $m$ the group index, $n$ the number of groups, $Loss_{m}$ the loss for group $m$, where a group is an ensemble member. 
At testing time, these groups will make predictions independently and their aggregated predictions are the final output for GENet. 
As a result of independent training and prediction, group heads function just like separate ConvNets.

As will be seen in experiments~(Table~\ref{tab:gain-ablation}), GENet is a non-trivial architecture design for explicit ensemble learning.
Through parameter sharing, it not only reduces the computation cost of standard ensemble methods but also gives an extra regularization for the shared parameters, since the representation learned from the shared-base is leveraged by multiple head classifiers.
In addition, the independent head classifiers provide sufficient diversity to group ensemble.

\begin{figure}[t!]
\centering
\includegraphics[width=.48\textwidth]{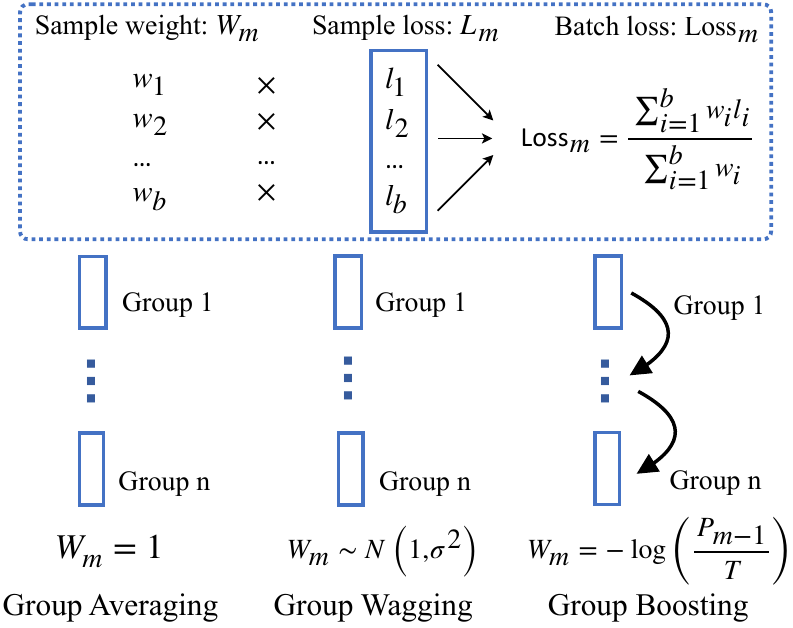}
\vspace{-1em}
\caption{\textbf{Model Aggregation Strategies.}
\textbf{Top:} Minibatch loss for each group (takes group  $m$ for demonstration here).
\textbf{Bottom:} Sample weight assignment mechanism.
\textbf{Group Averaging}: all samples are treated equally.
\textbf{Group Wagging}: sample weight randomly chosen from a distribution.
\textbf{Group Boosting}: sample weight is entropy of the former head classifier.
}
\vspace{-1.em}
\label{fig:archi}
\end{figure}

\subsection{Aggregation Strategies}
\label{sec:group-ens}
In order to ensemble multiple head classifiers into a final model, we investigate multiple aggregation strategies. 
Depending on the way we  assign weights to individual training samples~(illustrated in Figure~\ref{fig:archi}), these aggregation strategies can be divided into three types: posthoc aggregation (\eg., group averaging, stacking), independent aggregation~(\eg., group wagging), sequential aggregation~(\eg., group boosting). Minibatch loss for each head classifier is given by:
\begin{equation}
\label{equ:batch-loss}
    Loss_{m} = \frac{1} {\sum_{i=1}^{b} w_{i}} \sum_{i=1}^{b}w_{i}l_{i}
\end{equation}
where $Loss_{m}$ is the minibatch loss for group m, $i$ the sample index, $b$ the batch size, $ w_{i}$ the sample weight, $l_{i}$ the sample loss.

\paragraph{Group Averaging.}
For group averaging, all samples are treated equally without bias, which means $w_{i}$ fixed at 1 in Equation~\ref{equ:batch-loss}.
At testing time, their outputs are averaged as the final prediction.
The main idea behind group averaging is that due to different initialization, individual objective function, and separate gradient backpropagation, ensemble members will converge to different results.
Therefore, they will usually not make the same errors on the test set, reducing ensemble variance consequently.
To achieve ensemble diversity, these head classifiers can also be formed by using different architectures, objective functions, or training set splits.
A detailed analysis of ensemble output can be found in section~\ref{sec:ens-formulation-analy}.

\paragraph{Group Wagging.}
\label{sec:group-wag}
To introduce diversity among head classifiers, we propose group wagging, an aggregation method where each training sample is stochastically assigned a weight to mimic a perturbed training set. 
Sample weights can be drawn from many common distributions, \eg, uniform distribution, Gaussian distribution. 
When sampling from a Gaussian distribution, the instance weight is:
\begin{equation}
    w \sim N\left(1, \sigma^2\right),
\label{eq:wag-dis}
\end{equation}
where $w$ is the sample weight in Equation~\ref{equ:batch-loss}, $\sigma$ the standard deviation.
By changing $\sigma$, one can trade off bias and variance in group wagging. For example, when increasing the standard deviation $\sigma$, more samples will have their weights decrease to zero and disappear, thus increasing bias and reducing variance.
This is expected as different heads will have different convergence in such a case, thus leading to higher ensemble diversity.
Group averaging can be seen as a special case for group wagging when $\sigma$ is $0$, where all samples are treated equally. 

\paragraph{Group Boosting.}
\label{sec:group-boosting}
Besides group averaging and wagging, we also propose group boosting for dependent aggregation. Boosting is a general strategy to improve accuracy by repeatedly running a weak classifier on various distributed training samples. 
The next classifier will focus more on the misclassified samples when generating its training set.

As boosting is designed for sequential model training, the error rate and prediction results on the whole training set are used to construct the next classifier, which requires too many resources for deep ConvNets.
However, since all constituent members are training in parallel for group ensemble, we utilize an online boosting strategy per minibatch. 
Specifically, sample entropy of a head classifier is used as the weight for the next classifier,
\begin{equation} 
    W_{m} = \text{CrossEntropy}\left(\frac{P_{m-1}}{T}\right)
\label{eq:group-boost}
\end{equation}
where $W_{m}$ is sample weights for classifier $m$, 
$P_{m-1}$ is the logit output for classifier $m-1$,
T is temperature for softmax.
The minibatch loss is the same as in Equation~\ref{equ:batch-loss}, while the sample weights are replaced by the entropy of the former classifier. For hard samples of a classifier, the next classifier will focus more on them due to their high entropy.
At testing time, all head predictions are combined into the final one for GENet.

\begin{table*}[t!] 
\centering
\caption{\textbf{Accuracy Gains~(\%)} with different \textbf{GENet architectures} on CIFAR-100}
\vspace{0.5em}
\resizebox{.9\textwidth}{!}{
\begin{tabular}{@{}l|ccc|cccc|cccc@{}}
\toprule
 & \multicolumn{3}{c|}{Groups in ResNet-56}       & \multicolumn{4}{c|}{Groups in ResNet-110}       & \multicolumn{4}{c}{Groups in Wide ResNet-56-2}    \\
Split layer & 2& 3& 4& 2& 3& 4& 5& 2& 3& 4& 5       \\ \midrule
Conv2 & +1.79 & +1.78 & +1.5 & +1.65 & +2.12 & +1.7 & +1.38 & +1.22 & +1.42 & +2.11 & +1.65  \\
Conv3 & +1.78 & \textbf{+2.69} & +2.4 & +2.18 & +2.51 & \textbf{+2.79} & +2.49 & +1.63 & +2.22 & \textbf{+2.58} & +1.87 \\
Conv4 & +2.28 & +1.75 & +1.22 & +2.43 & +2.3 & +1.92 & +1.58 & +1.84 & +2.02 & +1.46 & +1.14  \\
FC & +0.89 & +0.74 & +0.93 & -0.12 & +0.16 & +0.58 & +1.54 & -0.09 & +0.73 & +1.03 & +1.37 \\
\bottomrule
\end{tabular}
}
\label{tab:expand-cifar-ablation}
\vspace{-0.5em}
\end{table*}

\subsection{Analysis of Ensemble Output} \label{sec:ens-formulation-analy}
Here we present an analysis to better understand the output of an ensemble (inspired by~\cite{Goodfellow-et-al-2016}). We consider the simple case of binary classification. Suppose we have a model ensemble where each models' output comes from the same normal distribution, 
\begin{equation} \label{eq:out}
y_{i} \sim N\left(\mu, \sigma^{2}\right), \quad \quad  i = 1,2,...,n
\end{equation}
where $y$ is the softmax output for the correct category, $i$ is the model index, $n$ the ensemble number,
$\mu$ \iffalse \footnote{For binary classification here, $\mu > 0.5.$} \fi the mean value and $\sigma$ the standard deviation.

First, we start with independent distributions, where the average output of the ensemble is also normally distributed,
\begin{equation} 
\label{eq:sum_out}
    \hat{y} = \frac{1}{n} \sum_{i=1}^{n} y_{i}, \quad
    \hat{y} \sim N\left(\mu, \frac{1}{n} \sigma^{2}\right)
\end{equation}
As shown by Equation~(\ref{eq:sum_out}), averaging outputs of multiple independent models reduces the variance, thus stabilizing the ensemble output.

Next, we consider the ensemble correlation. For two models with correlation $\rho$, the ensemble output is given by:
\begin{equation} \label{eq:sum_distribution}
    \hat{y} \sim N\left(\mu, \frac{1 + \rho}{2} \sigma^{2}\right).
\end{equation}
Due to the correlation factor $\rho$, the variance of ensemble output gets higher than independent members~($\rho = 0$).
However, as long as they are not perfectly correlated~($\rho = 1$), 
group ensemble will still outperform any of its members. In other words, group ensemble will perform at least as well as any of its members, and it will perform significantly better if the members are independent.

\begin{figure}[t!]
\centering
\includegraphics[width=\linewidth]{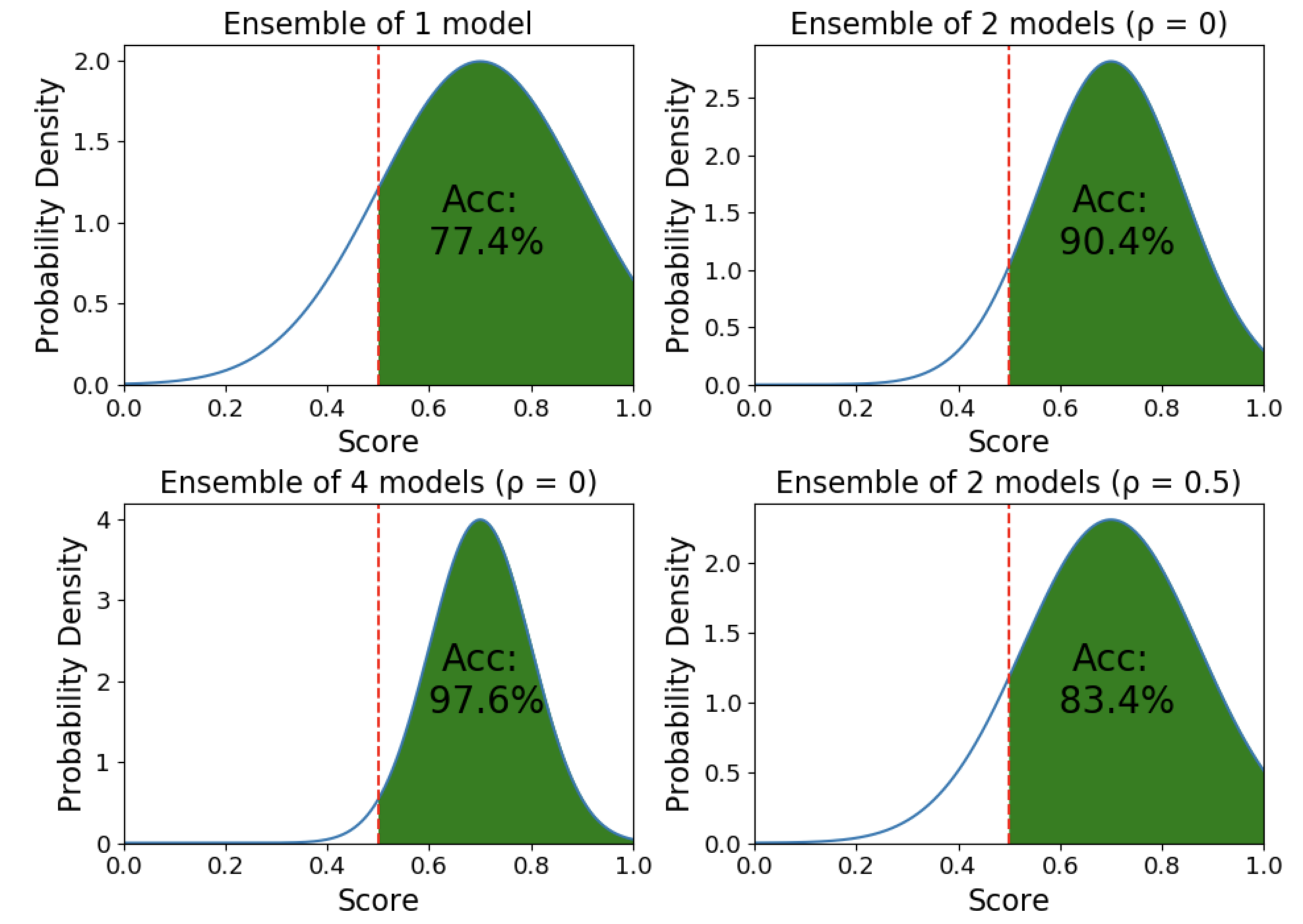}
\vspace{-.5em}
\caption{\textbf{Ensemble output distribution.}
\textbf{Top Left:} Baseline of a single model.
\textbf{Top Right:} Ensemble output of 2 independent models.
\textbf{Bottom Left:} Ensemble output of 4 independent models.
\textbf{Bottom Right:} Ensemble output of 2 models with correlation 0.5.
$\rho$ is the correlation among ensemble members.
}
\vspace{-1em}
\label{fig:ens-dis}
\end{figure}

Figure~\ref{fig:ens-dis} illustrates this analysis. It shows the ensemble output distribution for binary classification.
At a given capacity for constituent members, more groups and higher independence will generally lead to better performance.
Therefore, for group wagging, randomness and perturbation in minibatch gradient descent further reduces the correlation among groups, leading to higher accuracy as shown in Figure~\ref{fig:ens-dis} and Equation~\ref{eq:sum_distribution}. On the contrary, entropy boosting introduces high dependence among branches, which might deteriorate the ensemble diversity.

\section{Experiments on Image Recognition} 

\subsection{Datasets and Implementation Details}
\noindent{\textbf{CIFAR}} \cite{Krizhevsky2009LearningML} is an image recognition dataset with 50K training images and 10K validation images, 10 categories for CIFAR-10 and 100 categories for CIFAR-100.
ResNet-56~\cite{he2016deep}, ResNet-110, and Wide ResNet-56-2~(with 2x channels) with bottleneck block are used,
which start with a $3 \times 3$ convolution layer, followed by 3 stages, each with $n$ blocks (6 for ResNet-56, 12 for ResNet-110), and end with a fully-connected (FC) prediction layer.
For the rest of this paper, we refer to the first block of stage 1 as Conv2, and the same goes for other stages (\eg.,  Conv3 for the first block of stage 2).
% The filter numbers are 16, 32, and 64 for each stage, respectively.
Models are trained for 300 epochs with minibatch size 128, weight decay 0.0001, initial learning rate 0.25, decay by 0.1 at epoch 180 and 240,
and all other experiment settings follow~\cite{he2016deep}.
Top-1 error rates of center-crop are reported on validation set here.

\begin{figure}[t!]
\centering
\includegraphics[width=\linewidth]{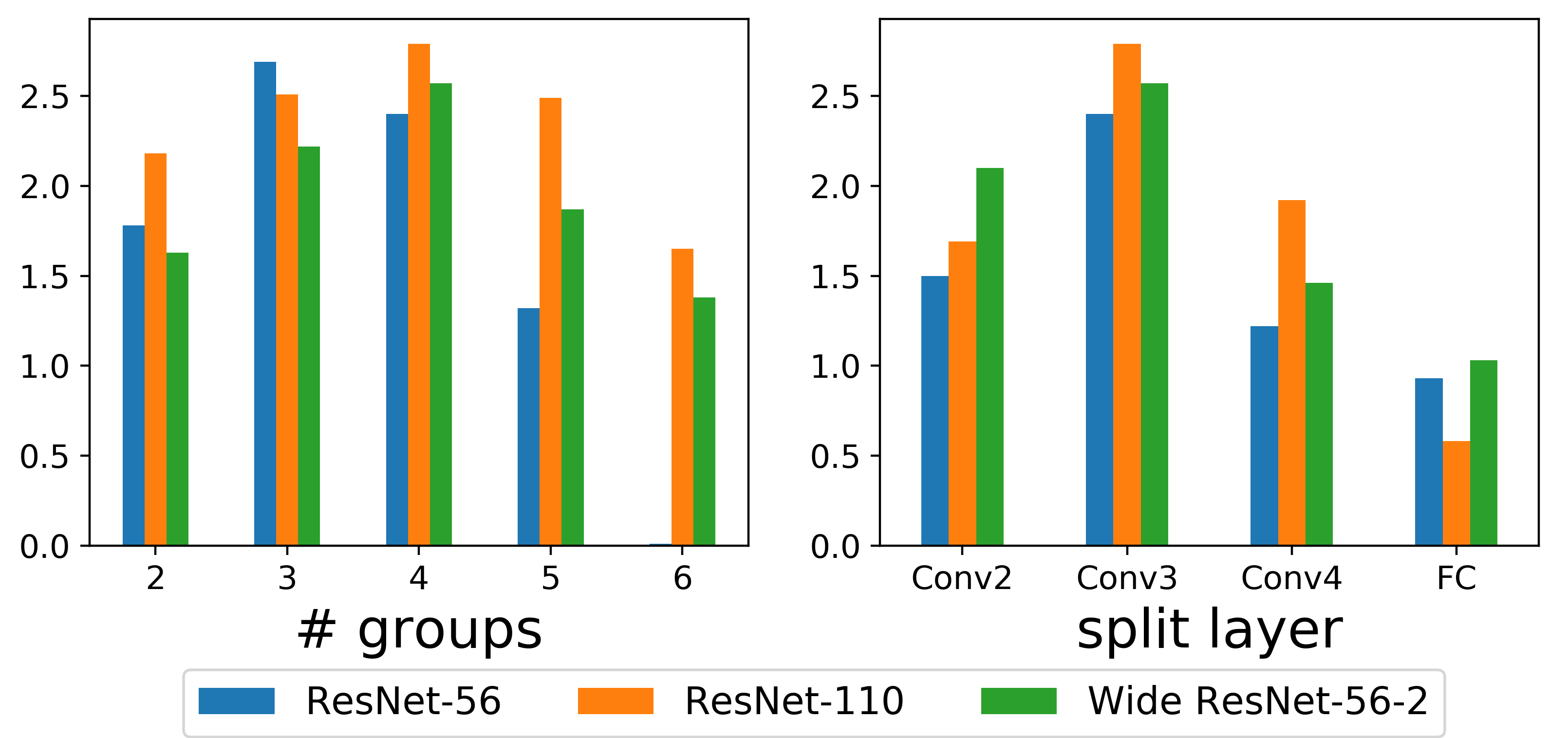}
\vspace{-0.5em}
\caption{\textbf{Ablation study on group number and split layer}.
Accuracy gain~(\%) on CIFAR-100 reported here.
\textbf{Left}: Splitting at layer Conv3.
\textbf{Right}: With 4 groups.
}
\vspace{-1em}
\label{fig:cifar-ablation}
\end{figure}

\noindent{\textbf{ImageNet}} consists of 1.2M training images and 50k validation images for 1K categories.
We train ResNet-50~\cite{he2016deep}, ResNeXt-50($32\times4$d)~\cite{xie2017aggregated} with batch size 512, initial learning rate 0.2 and a warm-up strategy as in~\cite{Goyal2017AccurateLM} for 300 epochs, decay by 0.1 at epoch 90, 180 and 240.
All other experiment settings follow~\cite{he2016deep}.
The $224 \times 224$ center crops are evaluated on the validation set.

In general, we will keep the parameters comparable when constructing GENet. For example, when we increase the number of groups (or heads), we will reduce the number of channels per-group. 
Standard deviation $\sigma$ is 0.2 in Equation~\ref{eq:wag-dis},
temperature $T$ is 2 in Equation~\ref{eq:group-boost},
and group averaging is the default aggregation strategy.
All experiments are implemented using PyTorch~\cite{Paszke2017AutomaticDI}.

\subsection{Ablation Study}
\noindent{\textbf{How many groups work best?}}
We first study the group numbers for GENet.
As shown in Figure~\ref{fig:cifar-ablation} left, when increasing groups from 2 to 6, the ensemble gain first increases and then decreases.
This is expected as more groups generally mean higher diversity for ensemble members but too many groups will reduce the head size and thus deteriorating its capacity.
Therefore, the most suitable group number is the one that balances well between member capacity and ensemble diversity.
When splitting at layer Conv3, 4 groups work best for ResNet-110 and wide ResNet-56-2, and 3 works best for ResNet-56. 
More results for the ablation study can be found in Table~\ref{tab:expand-cifar-ablation}.

\medskip
\noindent{\textbf{Where to split the network?}}
Then, we investigate the best split layer for GENet in Figure~\ref{fig:cifar-ablation} right.
Similar to the previous observation, the best split layer is the one that balances well between member capacity and their independence.
Although sharing more layers increases the member capacity, it also leads to a higher correlation between groups.
With 4 groups, splitting at layer Conv3 works best for all three models, ResNet-56, ResNet-110 and Wide ResNet-56-2.

\medskip
\noindent{\textbf{Group Averaging \vs Wagging \vs Boosting.}}
For different aggregation methods, results of posthoc combination~(group averaging), independent learning~(group wagging) and sequential learning~(group wagging) are shown in Table~\ref{tab:group-ens-ablation}, where
group wagging achieves the lowest errors on both CIFAR-100 and ImageNet.
It suggests that for group wagging, training perturbation enhances the ensemble diversity as expected, thus achieving the best performance.
On the contrary, by introducing dependence between ensemble members,
group boosting performs comparably or even worse with group averaging, which does not show any advantage as boosting does in standard sequential training.
This confirms the observation in~\cite{lee2015m,li2019ensemblenet} that ensemble-aware training often hurts for a multl-head network due to overfitting.
More analysis can be found in section~\ref{sec:ens-formulation-analy}.

\begin{table}[t!]
\centering
\caption{
\textbf{Aggregation strategies} for group ensemble.
Top-1 error~(\%) reported}
\vspace{0.5em}
% \resizebox{.48\textwidth}{!}{
\begin{tabular}{@{}l|cccc@{}}
\toprule
 Dataset & Groups & Averaging & Wagging        & Boosting       \\ \midrule
%  Dataset &groups & ave & wag       & boosting       \\ \hline
 &2       & 24.31   & \textbf{23.83} & 23.98          \\
CIFAR-100 &3          & 23.9    & \textbf{23.57} & 24.06          \\
&4           & 24.19   & \textbf{24.15} & 24.21 \\ 
\hline
ImageNet&2 & 22.12   & \textbf{21.81} & 22.24  \\ 
\bottomrule
\end{tabular}
% }
% \vspace{-1em}
\label{tab:group-ens-ablation}
\end{table}

\begin{table}[t!]
\centering
\caption{\textbf{Output Combination} for group ensemble.
Top-1 error~(\%) reported}
\vspace{0.5em}
% \resizebox{.48\textwidth}{!}{
\begin{tabular}{@{}l|ccc@{}}
\toprule
%  & CIFAR-10 & CIFAR-100 & ImageNet \\
 & CIFAR-10 & CIFAR-100 & ImageNet \\
\midrule
Probability & 4.12 & 23.96 & 21.92 \\
Logit & \textbf{4.1} & \textbf{23.57} & \textbf{21.81} \\
\bottomrule
\multicolumn{4}{c}{}\\
\multicolumn{4}{c}{}\\
\end{tabular}
% }
\label{tab:logit}
\vspace{-2em}
\end{table}

\begin{table}[t!]
\centering
\caption{Performance when \textbf{gradually adding GE components} on CIFAR-100.
Shared-base means parameter sharing,
ExplicitEns means explicit ensemble learning}
\vspace{0.5em}
\renewcommand{\arraystretch}{1.2}
\renewcommand{\tabcolsep}{1.4mm}
\resizebox{0.48\textwidth}{!}{
\begin{tabular}{@{}l|ccc@{}}
\toprule
 & ResNet-56 & ResNet-110 & Wide Res-56-2 \\
 \midrule
Baseline &29.25  &27.47  &26.13  \\
+ Shared-base & 28.61 (+0.64) & 26.24 (+1.23) & 24.98 (+1.15) \\
+ ExplicitEns & 23.9 (+4.71) & 21.71 (+4.53) & 20.61 (+4.37) \\
\bottomrule
\end{tabular}
}
\label{tab:gain-ablation}
\end{table}

\noindent{\textbf{Output combination.}}
Finally, for output combination, we try two different techniques at test time, averaging the logits~(outputs before softmax) or probability (outputs after softmax).
As shown in Table~\ref{tab:logit}, averaging logits is a better choice for combination in all cases.

\subsection{Experimental Analysis}
In this section, we explore GENet and its members in more depth to better understand how group ensemble works.

\begin{table}[t!]
\centering
\caption{\textbf{Model Ensemble \vs Group Ensemble} with similar parameters.
Top-1 error~(\%) on CIFAR-100
}
\vspace{0.5em}
\renewcommand{\arraystretch}{1.2}
\renewcommand{\tabcolsep}{1.4mm}
\resizebox{0.48\textwidth}{!}{

\begin{tabular}{@{}l|cc|cc@{}}

\toprule
 & \multicolumn{2}{c|}{ResNet-56} & \multicolumn{2}{c}{ResNet-110} \\
Groups & Model-Ens & Group-Ens & Model-Ens & Group-Ens \\ \midrule
2 & 25.8 (0.6M) & \textbf{23.8} (0.6M) & 24.0 (1.2M) & \textbf{22.0} (1.2M) \\
3 & 26.1 (0.6M) & \textbf{23.6} (0.6M) & 24.4 (1.2M) & \textbf{21.7} (1.2M) \\
4 & 25.5 (0.7M) & \textbf{24.1} (0.6M) & 23.9 (1.2M) & \textbf{21.6} (1.2M) \\
\bottomrule
\end{tabular}
}
\label{tab:Model-Group-Ensemble.}
% \vspace{-1em}
\end{table}

\medskip
\noindent{\textbf{Parameter Sharing}}
First, we present the results when gradually adding GENet components in Table~\ref{tab:gain-ablation}.
The baseline is training with one head, and the parameter sharing result is the average performance of all head classifiers.
As can be seen, parameter sharing not only reduces the computation cost but also improves the model capacity for ensemble members.
As analyzed in Section~\ref{sec:net-archi}, the feature representation learned in the shared-base is leveraged by multiple head classifiers, which adds an extra regularization to the shared parameters.
To show its effectiveness and fair comparison with model ensemble, we construct model ensemble and group ensemble with similar parameters~(Table~\ref{tab:Model-Group-Ensemble.}).
At given resources, group ensemble shows great advantage owing to the computational efficiency brought by parameter sharing.

\medskip
\noindent{\textbf{Explicit Ensemble Learning.}}
Then we analyze the contribution of explicit ensemble learning.
As can be seen in Table~\ref{tab:gain-ablation}, although ResNet utilizes implicit ensemble learning which can be interpreted as an ensemble of multiple shallow networks~\cite{veit2016residual}, it can still benefit greatly from explicit ensemble learning.
We note that it leads to 4.71\%/4.53\%/4\%.37 gain for ResNet-56/ResNet-110/Wide ResNet-56-2 respectively on CIFAR-100.
And due to the parameter sharing,  GENet can fully leverage the advantage of explicit ensemble learning while retaining the computation cost.

\begin{table}[t!]
\centering
\caption{GE with \textbf{implicit ensemble architecture}: ResNeXt~\cite{xie2017aggregated}.
 \newline   Top: CIFAR top-1 errors~(\%). 
 Bottom: ImageNet results}
\renewcommand{\arraystretch}{1.2}
\renewcommand{\tabcolsep}{1.4mm}
\vspace{0.5em}
\resizebox{0.48\textwidth}{!}{
\begin{tabular}{@{}l|cccc@{}}
\toprule
Methods & params & FLOPs & CIFAR10 & CIFAR100 \\
\midrule
ResNext-29-32x4d &  4.9M & 0.79G & 4.38 & 20.9 \\
+ Group Ensemble & 4.9M & 0.76G & \textbf{3.95} & \textbf{18.61}  \\
\bottomrule
% \vspace{.5em}
\toprule
Methods & params & FLOPs & err@1 & err@5  \\
\midrule
ResNext-50-32x4d & 25M & 4.29G & 22.13 & 6.17 \\
+ Group Ensemble & 23.9M & 4.22G & \textbf{20.3} & \textbf{5.3} \\
\bottomrule
\end{tabular}
}
\label{tab:rex-backbone}
\end{table}

\begin{table}[t!]
\centering
\caption{
Group ensemble with \textbf{implicit ensemble regularization} (Shake-Shake~\cite{Gastaldi17ShakeShake})
}
\vspace{0.5em}
\renewcommand{\arraystretch}{1.2}
\renewcommand{\tabcolsep}{1.4mm}
\resizebox{0.48\textwidth}{!}{

\begin{tabular}{@{}l|cccc@{}}
\toprule
Methods & params & FLOPs & CIFAR10 & CIFAR100 \\
\midrule
Shake-Shake-26-2x32d & 3M & 0.43G & 4.34 & 22.27 \\
+ Group Ensemble & 3M & 0.44G & \textbf{3.99} & \textbf{20.59} \\
\midrule
Shake-Shake-26-2x64d & 11.9M & 1.69G & 3.47 & 20.46 \\
+ Group Ensemble & 11.9M & 1.67G & \textbf{3.15} & \textbf{18.35} \\
\bottomrule
\end{tabular}
}

\label{tab:shake-ens}
% \end{minipage}
% \vspace{-1em}
\end{table}

\medskip
\noindent{\textbf{Combination with Implicit Ensemble Learning.}}
As discussed in Section~\ref{sec:intro}, implicit ensemble strategies are widely used in deep learning and can be divided into two types:
one is through `multiple paths' for architecture design~(\eg., ResNet~\cite{he2016deep}, ResNeXt~\cite{xie2017aggregated}, DenseNet~\cite{Huang2016DenselyCC}), 
and the other is through stochastic operations as a regularization strategy~(\eg, DropOut~\cite{Srivastava2014DropoutAS},
DropConnect~\cite{wan2013regularization}, StochDepth~\cite{huang2016deep},  Shape-Shake~\cite{Gastaldi17ShakeShake}).
To see if these approaches can be further enhanced by leveraging explicit ensemble learning, we conduct group ensemble based on ResNeXt~\cite{xie2017aggregated} and Shake-Shake~\cite{Gastaldi17ShakeShake}.

\emph{a) ResNeXt.} Similar to GENet, ResNeXt also utilizes group convolution to improve model capacity while retaining the computation cost.
As shown in Table~\ref{tab:rex-backbone}, although ResNeXt leverages implicit ensemble learning through dense group convolution and residual shortcut, it can benefit from group ensemble without extra computation cost, while fully leveraging the advantage of explicit ensemble learning.
A 1.83\% gain is seen on ImageNet with ResNeXt-50.

\emph{b) Shake-Shake.} Shake-Shake is an implicit ensemble regularization strategy which combines the output of multiple branches stochastically for the residual block at training time, and averaging their results for testing samples as an ensemble strategy.
As shown in Table~\ref{tab:shake-ens}, group ensemble improves the shake-shake baseline greatly while retaining the same computation of a single ConvNet.

\begin{figure}[t!]
\centering
\includegraphics[width=.48\textwidth]{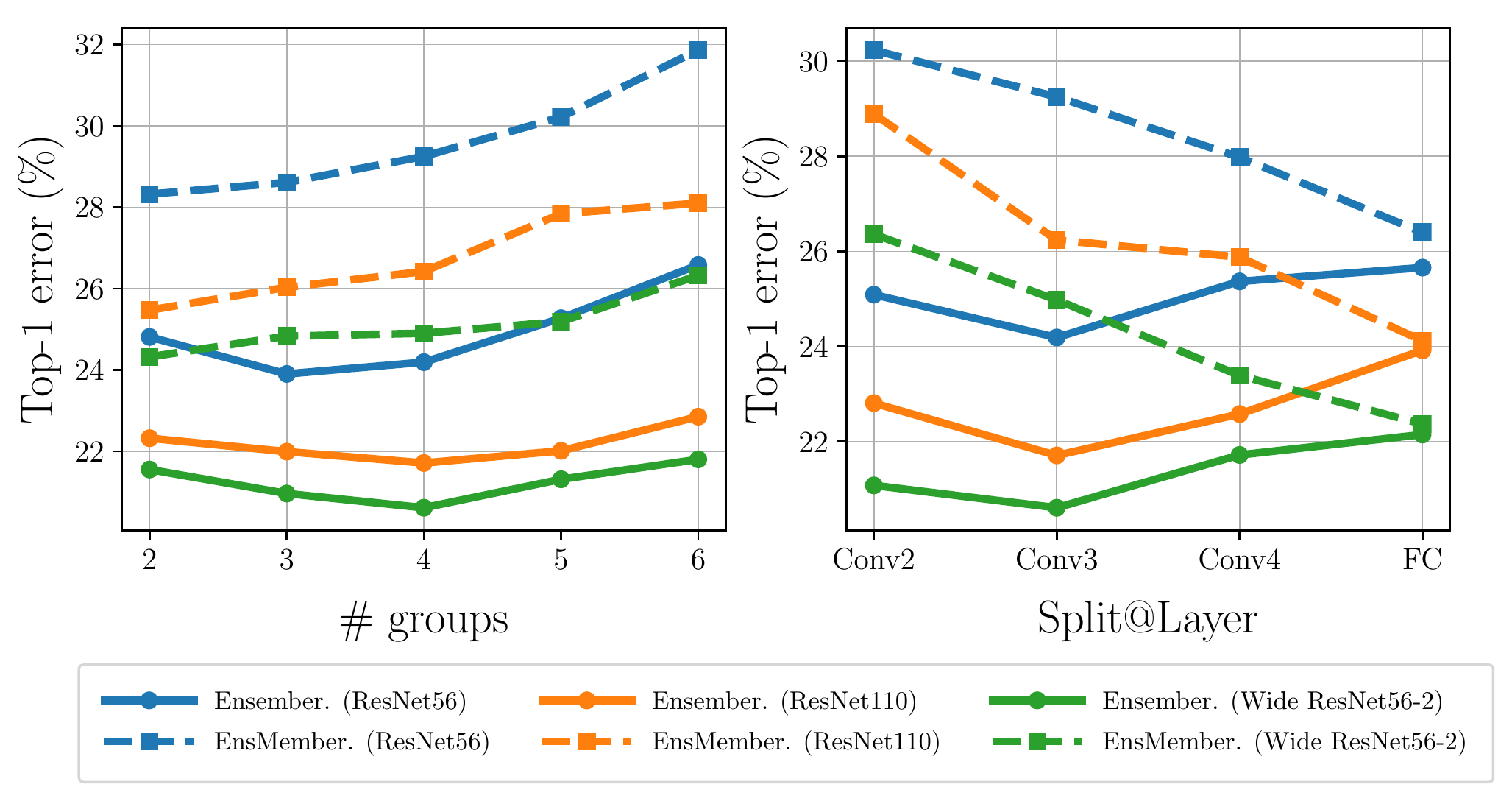}
\vspace{-1em}
\caption{\textbf{Group Ensemble \vs Ensemble Member} on CIFAR-100.
\textbf{Left:} Ensemble diversity~(gap between dashed and solid lines) is enhanced when adding more groups while members~(dashed lines) are getting weaker.
\textbf{Right:} When sharing more parameters, ensemble members are getting stronger at the cost of ensemble diversity.}
\label{fig:cifar-ens-member-gap}
\vspace{-.5em}
\end{figure}

\begin{figure}[t!]
    \centering
    \includegraphics[width=\linewidth]{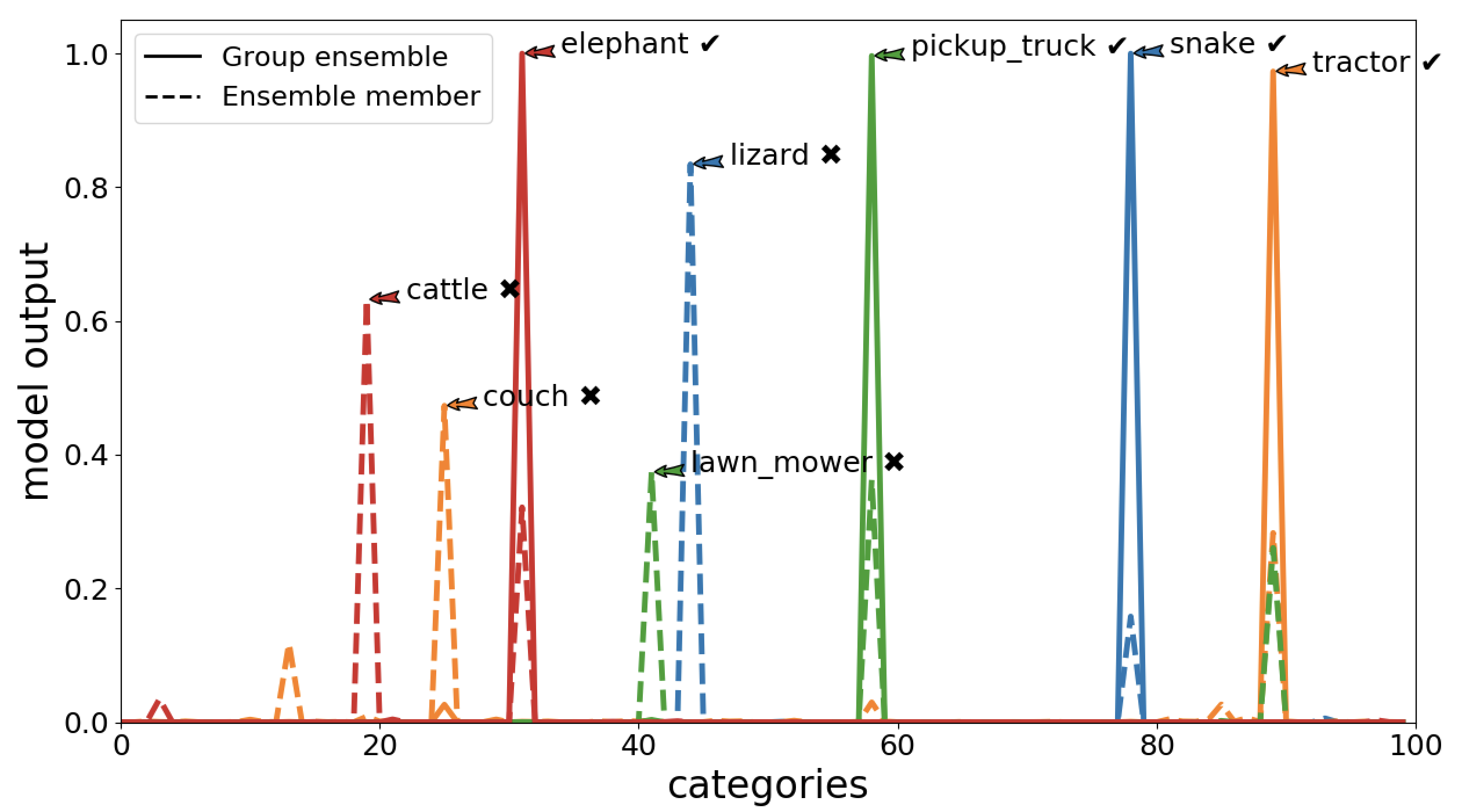}
    \vspace{-1em}
    \caption{\textbf{Prediction of Group Ensemble \vs Ensemble member} on CIFAR-100.
    Different colors correspond to different samples.
 For samples where a single model fails, GENet moves toward the correct category after the ensemble voting}
    \label{fig:ens-dist-gap}
    % \vspace{-0.5em}
\end{figure}
\medskip
\noindent{\textbf{Group Ensemble \vs Ensemble Member.}}
The performances of ensemble and its member are shown in Figure~\ref{fig:cifar-ens-member-gap}, where member performance is represented by their average error.
Their relationship shown in this figure also agrees with our observation in the ablation study.
In Figure~\ref{fig:cifar-ens-member-gap} left, adding more groups leads to larger ensemble diversity, which can be reflected by the performance gap between ensemble and its members.
However, too many groups will weaken  the head capacity and  deteriorate the final ensemble consequently.
% , which can be seen from the solid line.
For Figure~\ref{fig:cifar-ens-member-gap} right, splitting too early~(\eg, Conv2) hurt the member capacity while splitting too late~(\eg, FC) degenerates the ensemble diversity. 
This should be expected as ensemble members will converge to similar results when a large part of parameters are sharing, otherwise, the small model size will limit its capacity.

Figure~\ref{fig:ens-dist-gap} shows the prediction of group ensemble and its members.
As can be seen, for testing samples where a single model fails, GENet moves toward the correct category after the aggregation from different members.
It suggests that as long as the ensemble is constructed by diverse and varied models, group ensemble will perform better than any of its members.

\begin{figure}[t!]
    \centering
    \includegraphics[width=\linewidth]{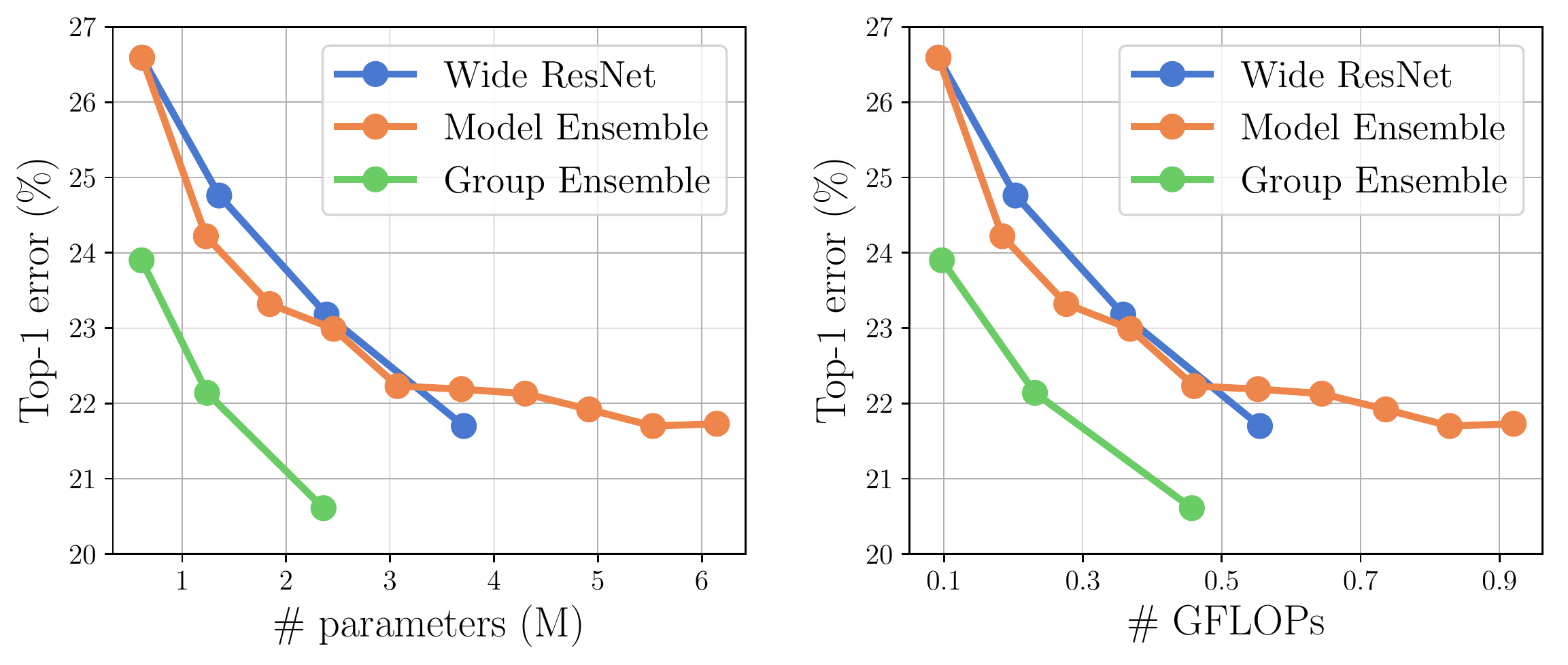}
    \vspace{-1em}
    \caption{\textbf{Computational efficiency} of GENet on CIFAR-100
    }
    \label{fig:model-comparel}
        \vspace{-.3em}
\end{figure}

\medskip
\noindent{\textbf{Computational Efficiency.}}
Finally, we investigate the computational efficiency of group ensemble in Figure~\ref{fig:model-comparel}.
To construct a series of model family, for model ensemble we simply add more independent models, while for group ensemble  we add filter width~(with best-fit groups).
And for a complete comparison, we also build the Wide ResNet-56 family as another baseline for increasing channel width.
As can be seen, increasing width for GENet is a much more effective way to build complex models.
An interesting observation is that a clear upper bound exists for standard model ensemble, while GENet can keep growing stronger by increasing groups and filter width.
Also, note that GENet outperforms Wide ResNet largely due to explicit ensemble learning.

\subsection{Comparisons with the State-of-the-art}

To compare with the state-of-the-art, we take approaches from two categories: other ensemble methods and novel architectures.
For ensemble methods, we choose standard model ensemble~\cite{Rokach2009EnsemblebasedC}, complementary embedding~\cite{Chen2019EmbeddingCD} and collaborative learning~\cite{song2018collaborative}.
Standard model ensemble~\cite{Rokach2009EnsemblebasedC} simply trains several independent models and average their predictions at testing time.
Complementary embedding \cite{Chen2019EmbeddingCD} trains multiple complementary neural networks sequentially, where each of them tackles the object categories in an easy-to-hard way by adapting the category importance to the error rates of the former classifier.
Collaborative learning~\cite{song2018collaborative} also utilizes the multi-head architecture which they refer to as intermediate-level representation~(ILR), together with a collaborative learning strategy for simultaneous head optimization.
Since GENet can be seen as a novel architecture, we also choose SENet~\cite{Hu_2018} and GCNet~\cite{cao2019GCNet} for comparison.

For CIFAR experiments in Table~\ref{tab:cifar-sot}, we follow the architecture design in~\cite{xie2017aggregated} where filter channels are 64, 128 and 256 for each stage respectively.
As can be seen, with similar parameters and FLOPs, group ensemble outperforms the baselines by a large margin.
Specifically, it improves the accuracy by 2.85\%/2.29\% for ResNet/ResNeXt backbone respectively on CIFAR-100.
Compared with the standard model ensemble, group ensemble achieves comparable performance while using only 30-40\% of computation resources~(in terms of parameters, FLOPs, and memory usage).

For ImageNet results in Table~\ref{tab:ilsvrc-sot}, group ensemble improves the top-1 error by 1.73\%/1.83\% respectively on the ResNet-50/ResNeXt-50 baseline.
Compared with the standard model ensemble, group ensemble achieves comparable performance using only 50\% of computation resources.
Additionally, GENet outperforms other ensemble approaches, \eg,  complementary embedding~\cite{Chen2019EmbeddingCD}, ILR~\cite{song2018collaborative} while consuming much fewer computation resources.
It shows that group ensemble is an effective architecture fully leveraging the advantage of explicit ensemble learning while retaining the computation cost.

\begin{table}[t!]
\caption{Comparison with the state-of-the-art on \textbf{CIFAR}}
\vspace{0.5em}
\centering
\renewcommand{\arraystretch}{1.2}
\renewcommand{\tabcolsep}{1.4mm}
\resizebox{0.48\textwidth}{!}{
\begin{tabular}{@{}l|cccc@{}}
\toprule
% % ResNet-34$^{*}$ & 21.3M & 3.78 & 20.99 \\
% IENet-34$^{*}$~\cite{gao2019intraensemble} & $>$22M & ~3.65 & - \\
% GENet-34$^{*}$ & 20.1M & \textbf{3.19} & \textbf{18.33} \\
% \midrule
Methods & params & FLOPS & CIFAR-10 & CIFAR-100 \\
 \midrule
ResNet-29 (baseline) & 5M & 0.76G & 5.24 & 22.98 \\
ResNet-29 Ensemble(2x) & 10M & 1.52G &4.77  &21.49  \\
ResNet-29 Ensemble(3x) & 15M & 2.28G &4.53  &\textbf{20.01}  \\
% ResNet-29 Ensemble(4x) & 20M & 3.04G &4.32  &19.53  \\
% ResNet-29 Ensemble(5x) & 25M & 3.8G &4.38  &19.21  \\
ResNet-29 + GE~(our) & 4.9M & 0.76G & \textbf{4.41} & 20.13 \\
\midrule
ResNext-29-32x4 (baseline) & 4.9M & 0.79G & 4.38 & 20.9 \\
ResNeXt29 Ensemble (2x) & 9.8M & 1.58G &4.02  &19.23  \\
ResNeXt29 Ensemble (3x)& 14.7M & 2.37G &\textbf{3.89}  &\textbf{18.17}  \\
% ResNeXt29 Ensemble (4x)& 19.6M & 3.16G &  &17.78  \\
ResNeXt29 + GE~(our) & 4.9M & 0.76G & 3.95 & 18.61 \\
\bottomrule
\end{tabular}
}
\label{tab:cifar-sot}
% \vspace{-1em}
\end{table}

\begin{table}[t!]
\centering
\caption{Comparison with the state-of-the-art on \textbf{ImageNet}}
\vspace{0.5em}
\renewcommand{\arraystretch}{1.2}
\renewcommand{\tabcolsep}{1.4mm}
\resizebox{0.48\textwidth}{!}{
\begin{tabular}{@{}l|cccc@{}}
\toprule
methods  &params. &FLOPs & err@1  &err@5 \\ \midrule
ResNet-50~\cite{he2016deep} & 25.6M &4.14G & 23.54 & 6.97 \\
ResNet-101~\cite{he2016deep} & 44.6M &7.85G & 21.87 & 6.29 \\
ResNet-50 + SE Layer~\cite{Hu_2018} & 28.1M &4.15G  & 23.29 & 6.62 \\
ResNet-50 + GC Layer~\cite{Hu_2018} & 28.1M &4.15G  & 22.3 & 6.34 \\
% \hline
% ResNet-50 + Dropout~\cite{Srivastava2014DropoutAS} & 25.6M & 23.2 & 6.59 \\
% % ResNet-50 + Cutout~\cite{devries2017cutout} & 25.6M & 22.93 & 6.66 \\
% % ResNet-50 + Mixup~\cite{Zhang2017mixupBE} & 25.6M & 22.58 & 6.4 \\
% ResNet-50 + StochDepth~\cite{huang2016deep} & 25.6M & 22.46 & 6.47 \\
% ResNet-50 + Manifold~\cite{Verma2018ManifoldMB} & 25.6M &22.50 &6.21 \\
% % ResNet-50 + AutoAugment~\cite{Cubuk2018AutoAugmentLA} & 25.6M &22.37 &6.18 \\
% ResNet-50 + DropBlock~\cite{Ghiasi2018DropBlockAR} & 51.2M & 21.87 &5.98 \\
% \hline
ResNet-50 + Embedding~\cite{Chen2019EmbeddingCD} & 102.4M &16.56G & 22.12 &6.79        \\
ResNet-50 + ILR~(2 heads)~\cite{song2018collaborative} & 49M &6.2G &22.7 &6.37 \\
ResNet-50 + ILR~(4 heads)~\cite{song2018collaborative} & 77.8M & 7.42G &22.29 &6.21 \\
ResNet-50 Ensemble (2x) & 51.2M &8.28G& 22.04 &6.11\\
ResNet-50 + GE~(our)  & 25.4M &4.21G  & \textbf{21.81} &\textbf{5.99}  \\ 
\midrule
ResNeXt-50~\cite{xie2017aggregated} & 25.0M   &4.29G   & 22.13   &6.17    \\
ResNeXt-50 + SE Layer~\cite{Hu_2018}  &27.5M &4.3G & 21.1 & 5.49 \\
ResNet-50 + ONE~\cite{zhu2018knowledge} & 39.58M &5.14G & 21.85 &5.9        \\
ResNeXt-50 Ensemble (2x) & 50M   &8.58G   & 20.54  &5.43    \\
ResNeXt-50 + GE~(our)   & 23.9M    &4.22G  & \textbf{20.3} &\textbf{5.3} \\ 
\midrule
ResNeXt-101~\cite{xie2017aggregated} & 44.2M   &8.03G   & 21.18   &5.57    \\
ResNeXt-101 + GE~(our)   & 44.9M    &8.35G  & \textbf{19.67} &\textbf{5.14} \\ 
\bottomrule
\end{tabular}
}
\label{tab:ilsvrc-sot}
\vspace{-1em}
\end{table}

% \vspace{-1em}
\section{Experiments on Action Recognition} \label{sec:sth_exp}
To show the effectiveness of group ensemble, we conduct experiments on action recognition with Something-Something~\cite{Goyal2017TheS} and Kinetics-200 dataset~\cite{Xie2017RethinkingSF} .

As a common practice, models are pre-trained on ImageNet and an inflation strategy is used for initialization \cite{Carreira_2017}.
For evaluation, we sample 10 clips for each video and average their predictions.
We take the ResNet-50 I3D model~\cite{Carreira_2017,Wang_2018} as the baseline, and 
8 frames are densely sampled from the original video on both datasets, all other settings follow the SlowFast Network~\cite{Feichtenhofer2018SlowFastNF}.

Table \ref{tab:sth} shows the results of group ensemble and other recent state-of-the-art approaches.  As can be seen, group ensemble improves the I3D baseline by 3\% on Something-Something and 1.4\% on Kinetics-200 dataset; and outperforms other state-of-the-art methods, reaching 50.7\%/80.1\% respectively. Similar to image recognition, group wagging performs better than group averaging owing to the training perturbations. 

\begin{table}[t]
\centering
\caption{\textbf{Action recognition accuracy (\%)}.
% Our group ensemble in based on I3D baseline, 
Our approach is based on I3D, where $ME$ means model ensemble, $GE_\text{avg}$ means groups averaging, $GE_\text{wag}$ means group wagging
}
\vspace{0.5em}
\renewcommand{\arraystretch}{1.2}
\renewcommand{\tabcolsep}{1.4mm}
\resizebox{0.48\textwidth}{!}{
\begin{tabular}{@{}l|ccc|cc@{}}
\toprule
% Datasets & I3D \cite{Wang2018VideosAS} & S3D-G\cite{Xie2017RethinkingSF} &  I3D+$GE_\text{avg}$ &  I3D+$GE_\text{wag}$  \\ 
Datasets & I3D  & S3D-G & I3D+$ME$&  I3D+$GE_\text{avg}$ &  I3D+$GE_\text{wag}$  \\ 
\midrule
Something$^2$ & 47.7 & 48.2 &50.1 & 49.2 & \textbf{50.7} \\ 
Kinetics-200 & 78.7 & 78.5 &80.0 & 79.8 & \textbf{80.1} \\ 
% Datasets & I3D\cite{Wang2018VideosAS} & ECO\cite{Zolfaghari2018ECOEC} & CorrNet\cite{wang2019video} & S3D-G\cite{Xie2017RethinkingSF}  & I3D + G  (our) \\ 
\bottomrule
\end{tabular}
}
\label{tab:sth}
\end{table}

\section{Experiments on Object Detection} \label{sec:det}
We further investigate group ensemble on object detection task using COCO 2017 dataset~\cite{lin2014microsoft}, which contains 118k training images and 5k validation images. We use Faster R-CNN~\cite{ren2015faster} based on ResNet-50 feature pyramid~\cite{Lin_2017} (following their basic implementation). As a two-stage object detection, Faster R-CNN is composed of two parts: region proposal network~(RPN) and R-CNN head.

For simplicity, we only utilize group ensemble in the R-CNN head while keeping the RPN unchanged. More complex combinations of group ensemble and Faster R-CNN will be explored for future work. As can be seen in Table~\ref{tab:det-coco}, group ensemble consistently improves the AP at different metrics~($AP, AP_{50}, AP_{75}$) and for different object sizes~($AP_{s}, AP_{m}, AP_{l}$). We note that it improves $AP_{75}$ by 1.54\%. It demonstrates that group ensemble is an effective and efficient way to improve model capacity across a wide range of common vision tasks.

\begin{table}[t]
\centering
\caption{\textbf{Object Detection} on COCO}
\vspace{0.5em}
\renewcommand{\arraystretch}{1.2}
\renewcommand{\tabcolsep}{1.4mm}
\resizebox{0.48\textwidth}{!}{
\begin{tabular}{@{}l|ccccccc@{}}
\toprule
 Methods  & params & AP & $AP_{50}$ & $AP_{75}$ & $AP_{s}$ & $AP_{m}$ & $AP_{l}$ \\ \midrule
Faster R-CNN & 41.7M & 38.2 & 59.2 & 41.2 & 21.7 & 41.8 & 49.3 \\
+ Model Ensemble & 83.4M & 38.9 & \textbf{60.1} & 42.1 & 23.2 & 42.3 & 50 \\
+ Group Ensemble & 41M & \textbf{39.2} & 60.0 & \textbf{42.7} & \textbf{23.2} & \textbf{42.5 }& \textbf{50.7} \\
\bottomrule
\end{tabular}
}
\label{tab:det-coco}
\vspace{-1em}
\end{table}

\section{Conclusion}
In this paper, we propose Group Ensemble Network~(GENet), which incorporates an ensemble of ConvNets in a single ConvNet. Through a shared-base and multi-head structure, it fully leverages the advantage of explicit ensemble learning while retaining the computation cost. In addition, we explore Group Averaging, Group Wagging, and Group Boosting to aggregate the constituent members in GENet. Finally, GENet outperforms larger single networks, standard model ensemble, and other recent state-of-the-art methods on CIFAR and ImageNet.  We also demonstrate its effectiveness on action recognition and object detection.

\paragraph{Acknowledgement.} This work was partially funded by DARPA MediFor
program under cooperative agreement FA87501620191.

{\small
\bibliographystyle{ieee_fullname}
\bibliography{egbib}

\begin{thebibliography}{10}\itemsep=-1pt

\bibitem{alpaydin1998techniques}
Ethem Alpaydin.
\newblock Techniques for combining multiple learners.
\newblock In {\em Proceedings of Engineering of Intelligent Systems}. Citeseer,
  1998.

\bibitem{Bauer1997AnEC}
Eric Bauer and Ron Kohavi.
\newblock An empirical comparison of voting classification algorithms: Bagging,
  boosting, and variants.
\newblock {\em Machine Learning}, 36:105--139, 1997.

\bibitem{bauer1999empirical}
Eric Bauer and Ron Kohavi.
\newblock An empirical comparison of voting classification algorithms: Bagging,
  boosting, and variants.
\newblock {\em Machine learning}, 36(1-2):105--139, 1999.

\bibitem{Breiman1996BaggingP}
Leo Breiman.
\newblock Bagging predictors.
\newblock {\em Machine Learning}, 24:123--140, 1996.

\bibitem{Breiman2001RandomF}
Leo Breiman.
\newblock Random forests.
\newblock {\em Machine Learning}, 45:5--32, 2001.

\bibitem{bromley1994signature}
Jane Bromley, Isabelle Guyon, Yann LeCun, Eduard S{\"a}ckinger, and Roopak
  Shah.
\newblock Signature verification using a" siamese" time delay neural network.
\newblock In {\em Advances in neural information processing systems}, pages
  737--744, 1994.

\bibitem{cao2019GCNet}
Yue Cao, Jiarui Xu, Stephen Lin, Fangyun Wei, and Han Hu.
\newblock Gcnet: Non-local networks meet squeeze-excitation networks and
  beyond.
\newblock {\em arXiv preprint arXiv:1904.11492}, 2019.

\bibitem{Carreira_2017}
Joao Carreira and Andrew Zisserman.
\newblock Quo vadis, action recognition? a new model and the kinetics dataset.
\newblock {\em 2017 IEEE Conference on Computer Vision and Pattern Recognition
  (CVPR)}, Jul 2017.

\bibitem{caruana1997multitask}
Rich Caruana.
\newblock Multitask learning.
\newblock {\em Machine learning}, 28(1):41--75, 1997.

\bibitem{Caruana2004EnsembleSF}
Rich Caruana, Alexandru Niculescu-Mizil, Geoff Crew, and Alex Ksikes.
\newblock Ensemble selection from libraries of models.
\newblock In {\em ICML}, 2004.

\bibitem{Chen2019EmbeddingCD}
Qiuyu Chen, Wei Zhang, Jun Yu, and Jianping Fan.
\newblock Embedding complementary deep networks for image classification.
\newblock In {\em CVPR}, 2019.

\bibitem{chollet2017xception}
Fran{\c{c}}ois Chollet.
\newblock Xception: Deep learning with depthwise separable convolutions.
\newblock In {\em CVPR}, pages 1251--1258, 2017.

\bibitem{cortes2014deep}
Corinna Cortes, Mehryar Mohri, and Umar Syed.
\newblock Deep boosting.
\newblock In {\em ICML}, 2014.

\bibitem{drucker1993improving}
Harris Drucker, Robert Schapire, and Patrice Simard.
\newblock Improving performance in neural networks using a boosting algorithm.
\newblock In {\em Advances in neural information processing systems}, pages
  42--49, 1993.

\bibitem{Feichtenhofer2018SlowFastNF}
Christoph Feichtenhofer, Haoqi Fan, Jitendra Malik, and Kaiming He.
\newblock Slowfast networks for video recognition.
\newblock {\em ArXiv}, abs/1812.03982, 2018.

\bibitem{Freund1996ExperimentsWA}
Yoav Freund and Robert~E. Schapire.
\newblock Experiments with a new boosting algorithm.
\newblock In {\em ICML}, 1996.

\bibitem{friedman2001elements}
Jerome Friedman, Trevor Hastie, and Robert Tibshirani.
\newblock {\em The elements of statistical learning}, volume~1.
\newblock Springer series in statistics New York, 2001.

\bibitem{Gastaldi17ShakeShake}
Xavier Gastaldi.
\newblock Shake-shake regularization.
\newblock {\em arXiv preprint arXiv:1705.07485}, 2017.

\bibitem{Ghiasi2018DropBlockAR}
Golnaz Ghiasi, Tsung-Yi Lin, and Quoc~V. Le.
\newblock Dropblock: A regularization method for convolutional networks.
\newblock In {\em NeurIPS}, 2018.

\bibitem{Gkioxari2014RCNNsFP}
Georgia Gkioxari, Bharath Hariharan, Ross~B. Girshick, and Jitendra Malik.
\newblock R-cnns for pose estimation and action detection.
\newblock {\em ArXiv}, abs/1406.5212, 2014.

\bibitem{Goodfellow-et-al-2016}
Ian Goodfellow, Yoshua Bengio, and Aaron Courville.
\newblock {\em Deep Learning}.
\newblock MIT Press, 2016.
\newblock \url{http://www.deeplearningbook.org}.

\bibitem{Goyal2017AccurateLM}
Priya Goyal, Piotr Doll{\'a}r, Ross~B. Girshick, Pieter Noordhuis, Lukasz
  Wesolowski, Aapo Kyrola, Andrew Tulloch, Yangqing Jia, and Kaiming He.
\newblock Accurate, large minibatch sgd: Training imagenet in 1 hour.
\newblock {\em ArXiv}, abs/1706.02677, 2017.

\bibitem{Goyal2017TheS}
Raghav Goyal, Samira~Ebrahimi Kahou, Vincent Michalski, Joanna Materzynska,
  Susanne Westphal, Heuna Kim, Valentin Haenel, Ingo Fr{\"u}nd, Peter Yianilos,
  Moritz Mueller-Freitag, Florian Hoppe, Christian Thurau, Ingo Bax, and Roland
  Memisevic.
\newblock The “something something” video database for learning and
  evaluating visual common sense.
\newblock {\em 2017 IEEE International Conference on Computer Vision (ICCV)},
  pages 5843--5851, 2017.

\bibitem{He_2017}
Kaiming He, Georgia Gkioxari, Piotr Dollar, and Ross Girshick.
\newblock Mask r-cnn.
\newblock {\em 2017 IEEE International Conference on Computer Vision (ICCV)},
  Oct 2017.

\bibitem{he2016deep}
Kaiming He, Xiangyu Zhang, Shaoqing Ren, and Jian Sun.
\newblock Deep residual learning for image recognition.
\newblock In {\em CVPR}, pages 770--778, 2016.

\bibitem{howard2017mobilenets}
Andrew~G Howard, Menglong Zhu, Bo Chen, Dmitry Kalenichenko, Weijun Wang,
  Tobias Weyand, Marco Andreetto, and Hartwig Adam.
\newblock Mobilenets: Efficient convolutional neural networks for mobile vision
  applications.
\newblock {\em arXiv preprint arXiv:1704.04861}, 2017.

\bibitem{Hu_2018}
Jie Hu, Li Shen, and Gang Sun.
\newblock Squeeze-and-excitation networks.
\newblock {\em 2018 IEEE/CVF Conference on Computer Vision and Pattern
  Recognition}, Jun 2018.

\bibitem{Huang2016DenselyCC}
Gao Huang, Zhuang Liu, Kilian~Q. Weinberger, and Laurens van~der Maaten.
\newblock Densely connected convolutional networks.
\newblock {\em 2017 IEEE Conference on Computer Vision and Pattern Recognition
  (CVPR)}, pages 2261--2269, 2016.

\bibitem{huang2016deep}
Gao Huang, Yu Sun, Zhuang Liu, Daniel Sedra, and Kilian~Q Weinberger.
\newblock Deep networks with stochastic depth.
\newblock In {\em European conference on computer vision}, pages 646--661.
  Springer, 2016.

\bibitem{kittler1998combining}
Josef Kittler, Mohamad Hatef, Robert~PW Duin, and Jiri Matas.
\newblock On combining classifiers.
\newblock {\em IEEE transactions on pattern analysis and machine intelligence},
  20(3):226--239, 1998.

\bibitem{Kornblith2019SimilarityON}
Simon Kornblith, Mohammad Norouzi, Honglak Lee, and Geoffrey~E. Hinton.
\newblock Similarity of neural network representations revisited.
\newblock {\em ArXiv}, abs/1905.00414, 2019.

\bibitem{Krizhevsky2009LearningML}
Alex Krizhevsky.
\newblock Learning multiple layers of features from tiny images.
\newblock 2009.

\bibitem{krizhevsky2012imagenet}
Alex Krizhevsky, Ilya Sutskever, and Geoffrey~E Hinton.
\newblock Imagenet classification with deep convolutional neural networks.
\newblock In {\em NIPS}, pages 1097--1105, 2012.

\bibitem{kuznetsov2014multi}
Vitaly Kuznetsov, Mehryar Mohri, and Umar Syed.
\newblock Multi-class deep boosting.
\newblock In {\em Advances in Neural Information Processing Systems}, pages
  2501--2509, 2014.

\bibitem{lee2015m}
Stefan Lee, Senthil Purushwalkam, Michael Cogswell, David Crandall, and Dhruv
  Batra.
\newblock Why m heads are better than one: Training a diverse ensemble of deep
  networks.
\newblock {\em arXiv preprint arXiv:1511.06314}, 2015.

\bibitem{li2019ensemblenet}
Hanhan Li, Joe Yue-Hei Ng, and Paul Natsev.
\newblock Ensemblenet: End-to-end optimization of multi-headed models, 2019.

\bibitem{Lin_2017}
Tsung-Yi Lin, Piotr Dollar, Ross Girshick, Kaiming He, Bharath Hariharan, and
  Serge Belongie.
\newblock Feature pyramid networks for object detection.
\newblock {\em 2017 IEEE Conference on Computer Vision and Pattern Recognition
  (CVPR)}, Jul 2017.

\bibitem{lin2014microsoft}
Tsung-Yi Lin, Michael Maire, Serge Belongie, James Hays, Pietro Perona, Deva
  Ramanan, Piotr Doll{\'a}r, and C~Lawrence Zitnick.
\newblock Microsoft coco: Common objects in context.
\newblock In {\em European conference on computer vision}, pages 740--755.
  Springer, 2014.

\bibitem{long2015learning}
Mingsheng Long and Jianmin Wang.
\newblock Learning multiple tasks with deep relationship networks.
\newblock {\em NIPS}, 2017.

\bibitem{meyerson2017beyond}
Elliot Meyerson and Risto Miikkulainen.
\newblock Beyond shared hierarchies: Deep multitask learning through soft layer
  ordering.
\newblock {\em ICLR}, 2018.

\bibitem{misra2016cross}
Ishan Misra, Abhinav Shrivastava, Abhinav Gupta, and Martial Hebert.
\newblock Cross-stitch networks for multi-task learning.
\newblock In {\em CVPR}, pages 3994--4003, 2016.

\bibitem{moghimi2016boosted}
Mohammad Moghimi, Serge~J Belongie, Mohammad~J Saberian, Jian Yang, Nuno
  Vasconcelos, and Li-Jia Li.
\newblock Boosted convolutional neural networks.
\newblock In {\em BMVC}, pages 24--1, 2016.

\bibitem{Morcos2018InsightsOR}
Ari~S. Morcos, Maithra Raghu, and Samy Bengio.
\newblock Insights on representational similarity in neural networks with
  canonical correlation.
\newblock In {\em NeurIPS}, 2018.

\bibitem{Paszke2017AutomaticDI}
Adam Paszke, Sam Gross, Soumith Chintala, Gregory Chanan, Edward Yang, Zachary
  Devito, Zeming Lin, Alban Desmaison, Luca Antiga, and Adam Lerer.
\newblock Automatic differentiation in pytorch.
\newblock 2017.

\bibitem{Peng_2016}
Zhanglin Peng, Ya Li, Zhaoquan Cai, and Liang Lin.
\newblock Deep boosting: Joint feature selection and analysis dictionary
  learning in hierarchy.
\newblock {\em Neurocomputing}, 178:36–45, Feb 2016.

\bibitem{perrone1994general}
Michael~P Perrone.
\newblock General averaging results for convex optimization.
\newblock In {\em Proceedings of the 1993 Connectionist Models Summer School},
  pages 364--371, 1994.

\bibitem{Pham2018EfficientNA}
Hieu Pham, Melody~Y. Guan, Barret Zoph, Quoc~V. Le, and Jeff Dean.
\newblock Efficient neural architecture search via parameter sharing.
\newblock {\em ArXiv}, abs/1802.03268, 2018.

\bibitem{Raghu2017SVCCASV}
Maithra Raghu, Justin Gilmer, Jason Yosinski, and Jascha Sohl-Dickstein.
\newblock Svcca: Singular vector canonical correlation analysis for deep
  understanding and improvement.
\newblock {\em NIPS}, abs/1706.05806, 2017.

\bibitem{ren2015faster}
Shaoqing Ren, Kaiming He, Ross Girshick, and Jian Sun.
\newblock Faster r-cnn: Towards real-time object detection with region proposal
  networks.
\newblock In {\em Advances in neural information processing systems}, pages
  91--99, 2015.

\bibitem{Rokach2009EnsemblebasedC}
Lior Rokach.
\newblock Ensemble-based classifiers.
\newblock {\em Artificial Intelligence Review}, 33:1--39, 2009.

\bibitem{rosenbaum2017routing}
Clemens Rosenbaum, Tim Klinger, and Matthew Riemer.
\newblock Routing networks: Adaptive selection of non-linear functions for
  multi-task learning.
\newblock {\em ICLR}, 2018.

\bibitem{ruder122019latent}
Sebastian Ruder12, Joachim Bingel, Isabelle Augenstein, and Anders S{\o}gaard.
\newblock Latent multi-task architecture learning.
\newblock 2019.

\bibitem{Russakovsky_2015}
Olga Russakovsky, Jia Deng, Hao Su, Jonathan Krause, Sanjeev Satheesh, Sean Ma,
  Zhiheng Huang, Andrej Karpathy, Aditya Khosla, Michael Bernstein, and et al.
\newblock Imagenet large scale visual recognition challenge.
\newblock {\em International Journal of Computer Vision}, 115(3):211–252, Apr
  2015.

\bibitem{schwenk1997adaboosting}
Holger Schwenk and Yoshua Bengio.
\newblock Adaboosting neural networks: Application to on-line character
  recognition.
\newblock In {\em International Conference on Artificial Neural Networks},
  pages 967--972. Springer, 1997.

\bibitem{schwenk2000boosting}
Holger Schwenk and Yoshua Bengio.
\newblock Boosting neural networks.
\newblock {\em Neural computation}, 12(8):1869--1887, 2000.

\bibitem{song2018collaborative}
Guocong Song and Wei Chai.
\newblock Collaborative learning for deep neural networks.
\newblock In {\em Advances in Neural Information Processing Systems}, pages
  1832--1841, 2018.

\bibitem{Srivastava2014DropoutAS}
Nitish Srivastava, Geoffrey~E. Hinton, Alex Krizhevsky, Ilya Sutskever, and
  Ruslan Salakhutdinov.
\newblock Dropout: a simple way to prevent neural networks from overfitting.
\newblock {\em J. Mach. Learn. Res.}, 15:1929--1958, 2014.

\bibitem{Szegedy2016Inceptionv4IA}
Christian Szegedy, Sergey Ioffe, Vincent Vanhoucke, and Alex Alemi.
\newblock Inception-v4, inception-resnet and the impact of residual connections
  on learning.
\newblock In {\em AAAI}, 2016.

\bibitem{Szegedy_2015}
Christian Szegedy, Wei Liu, Yangqing Jia, Pierre Sermanet, Scott Reed, Dragomir
  Anguelov, Dumitru Erhan, Vincent Vanhoucke, and Andrew Rabinovich.
\newblock Going deeper with convolutions.
\newblock {\em 2015 IEEE Conference on Computer Vision and Pattern Recognition
  (CVPR)}, Jun 2015.

\bibitem{Szegedy_2016}
Christian Szegedy, Vincent Vanhoucke, Sergey Ioffe, Jon Shlens, and Zbigniew
  Wojna.
\newblock Rethinking the inception architecture for computer vision.
\newblock {\em 2016 IEEE Conference on Computer Vision and Pattern Recognition
  (CVPR)}, Jun 2016.

\bibitem{Tolstikhin2017AdaGANBG}
Ilya~O. Tolstikhin, Sylvain Gelly, Olivier Bousquet, Carl-Johann Simon-Gabriel,
  and Bernhard Sch{\"o}lkopf.
\newblock Adagan: Boosting generative models.
\newblock In {\em NIPS}, 2017.

\bibitem{Tran2019VideoCW}
Du Tran, Heng Wang, Lorenzo Torresani, and Matt Feiszli.
\newblock Video classification with channel-separated convolutional networks.
\newblock {\em ICCV}, 2019.

\bibitem{veit2016residual}
Andreas Veit, Michael~J Wilber, and Serge Belongie.
\newblock Residual networks behave like ensembles of relatively shallow
  networks.
\newblock In {\em Advances in neural information processing systems}, pages
  550--558, 2016.

\bibitem{wan2013regularization}
Li Wan, Matthew Zeiler, Sixin Zhang, Yann Le~Cun, and Rob Fergus.
\newblock Regularization of neural networks using dropconnect.
\newblock In {\em International conference on machine learning}, pages
  1058--1066, 2013.

\bibitem{Wang_2018}
Xiaolong Wang, Ross Girshick, Abhinav Gupta, and Kaiming He.
\newblock Non-local neural networks.
\newblock {\em 2018 IEEE/CVF Conference on Computer Vision and Pattern
  Recognition}, Jun 2018.

\bibitem{xie2017aggregated}
Saining Xie, Ross Girshick, Piotr Doll{\'a}r, Zhuowen Tu, and Kaiming He.
\newblock Aggregated residual transformations for deep neural networks.
\newblock In {\em CVPR}, pages 1492--1500, 2017.

\bibitem{Xie2017RethinkingSF}
Saining Xie, Chen Sun, Jonathan Huang, Zhuowen Tu, and Kevin Murphy.
\newblock Rethinking spatiotemporal feature learning: Speed-accuracy trade-offs
  in video classification.
\newblock In {\em ECCV}, 2017.

\bibitem{zhou2012ensemble}
Zhi-Hua Zhou.
\newblock {\em Ensemble methods: foundations and algorithms}.
\newblock Chapman and Hall/CRC, 2012.

\bibitem{zhu2018knowledge}
Xiatian Zhu, Shaogang Gong, et~al.
\newblock Knowledge distillation by on-the-fly native ensemble.
\newblock In {\em Advances in neural information processing systems}, pages
  7517--7527, 2018.

\end{thebibliography}
}

\end{document}